\documentclass[letterpaper, 10 pt, conference]{ieeeconf}  

\IEEEoverridecommandlockouts                              

\usepackage{graphics} 
\usepackage{epsfig} 
\usepackage{amsmath} 
\usepackage[english]{babel}
\usepackage[utf8]{inputenc}
\usepackage{algorithm}
\usepackage{siunitx}
\usepackage{color}
\usepackage{colortbl}
\usepackage{comment}
\usepackage{listings}
\usepackage{caption}
\usepackage{wrapfig,lipsum,booktabs}
\usepackage{booktabs}       
\usepackage[noend]{algpseudocode}

\title{\LARGE \bf
MaP-AVR: A Meta-Action Planner for Agents Leveraging Vision Language Models and Retrieval-Augmented Generation
}

\author{Zhenglong Guo*, Yiming Zhao*, Feng Jiang, Heng Jin, \\ Zongbao Feng, Jianbin Zhou, Siyuan Xu 
	\thanks{
		-----------------------------------------------------------------------------------------
		\newline 
		* Equal contribution. 
		\newline 
		All authors are affiliated with Li Auto.
		\newline
		{Email: zhaoyiming3@lixiang.com}
	}%
}
\begin{document}

\maketitle
\thispagestyle{empty}
\pagestyle{empty}


\begin{abstract}

Embodied robotic AI systems designed to manage complex daily tasks rely on a task planner to understand and decompose high-level tasks. While most research focuses on enhancing the task-understanding abilities of LLMs/VLMs through fine-tuning or chain-of-thought prompting, this paper argues that defining the planned skill set is equally crucial. To handle the complexity of daily environments, the skill set should possess a high degree of generalization ability. Empirically, more abstract expressions tend to be more generalizable. Therefore, we propose to abstract the planned result as a set of meta-actions. Each meta-action comprises three components: {move/rotate, end-effector status change, relationship with the environment}. This abstraction replaces human-centric concepts, such as grasping or pushing, with the robot's intrinsic functionalities. As a result, the planned outcomes align seamlessly with the complete range of actions that the robot is capable of performing. Furthermore, to ensure that the LLM/VLM accurately produces the desired meta-action format, we employ the Retrieval-Augmented Generation (RAG) technique, which leverages a database of human-annotated planning demonstrations to facilitate in-context learning. As the system successfully completes more tasks, the database will self-augment to continue supporting diversity. The meta-action set and its integration with RAG are two novel contributions of our planner, denoted as MaP-AVR—the meta-action planner for agents composed of VLM and RAG. To validate its efficacy, we design experiments using GPT-4o as the pre-trained LLM/VLM model and OmniGibson as our robotic platform. Our approach demonstrates promising performance compared to the current state-of-the-art method. Project page: \textit{https://map-avr.github.io/}


\end{abstract}

\section{INTRODUCTION} \label{intro}


A core capability of intelligent robots designed to assist humans in daily life is the ability to simultaneously understand both human needs and 3D environments. The emergence of large language models, particularly multimodal models such as GPT-4V and GPT-4o, showcases remarkable intelligence and has ignited hope for developing such advanced robots \cite{kim2024openvla, wake2024gpt, achiam2023gpt}. 

\begin{figure}
\includegraphics[width=1.\linewidth, height=50mm]{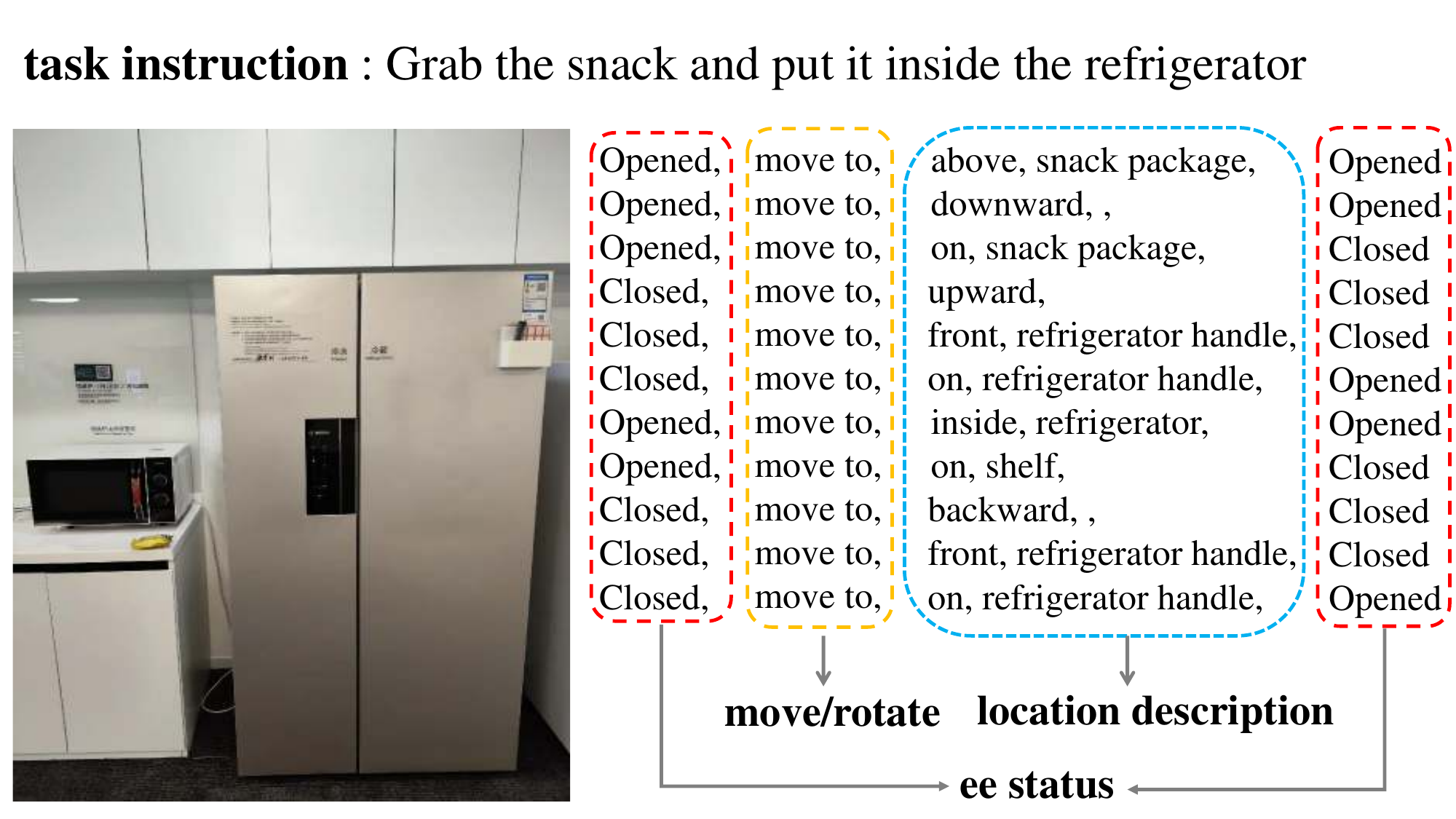}
    \caption{An example of a set of meta-actions as the planned outcome. Each meta-action comprises three essential components, which are utilized in subsequent execution functions.}
    \label{fig:meta_action}
    \vspace{-4mm}
\end{figure}

The optimal design of such intelligent robotic systems continues to be an open and active area of research. Incorporating language signals into models, which enable networks originally trained via reinforcement learning or imitation learning for single-task settings to handle multiple skills based on varying language inputs, has garnered significant research interest \cite{chi2023diffusion, ke20243d, brohan2022rt, goyal2023rvt, goyal2024rvt}. These employed language signals are typically short action commands rather than natural language task descriptions. To address this limitation, instead of relying on interactive approaches with humans \cite{brohan2023can}, recent research has shifted towards utilizing multimodal large language models as task planners responsible for decomposing high-level daily task descriptions into sequences of short action instructions \cite{mu2024embodiedgpt, driess2023palm}. Typically, these short action instructions are skills, such as grasping, placing, pushing, pulling, moving, and releasing, defined by human experience. However, whether these predefined human-centric skill sets can fully capture the vast diversity of daily tasks that the robot is able to complete remains an open question, raising concerns about the generalization ability of this type of skill set definition.

\textbf{Motivation} To address the aforementioned concerns, we propose that the planned results be structured as a set of meta-actions. Our idea is inspired by Richard S. Sutton's blog titled “The Bitter Lesson” \cite{exampleWebsite}, where he argues that AI agents should be constructed using only meta-methods capable of discovering and capturing arbitrary complexity, rather than relying on built-in human-centric knowledge. These high-capacity meta-methods are general-purpose and have the potential to approximate arbitrary complexity, thereby working towards achieving his vision: “We want AI agents that can discover like we can, not contain what we have discovered.” In light of this, we designed a meta-action planner, where the planned meta-actions represent the operations that the robot itself can inherently perform, allowing for a versatile representation of daily tasks through arbitrary combinations.

In 3D space, almost all robots possess the fundamental capabilities of translation and rotation. A robot's end-effector might be equipped with various tools or attachments, which
\begin{wrapfigure}{l}{.45\linewidth}
        \includegraphics[width=1.\linewidth, height=100mm]{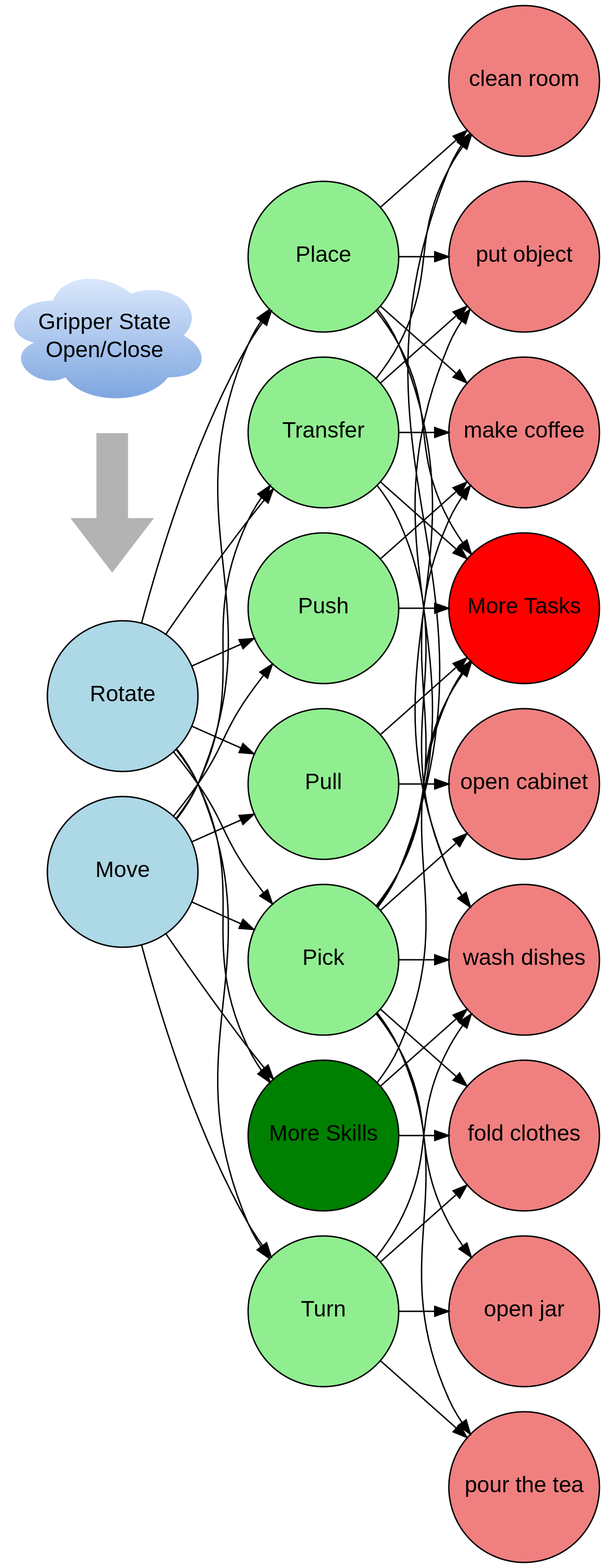}        
     \vspace{-5mm}
    \caption{The figure illustrates that meta-actions represent the foundational abstraction for many skills aligned with human experience. The flexible combinations of meta-actions can be used to compose a wide range of daily tasks.}\label{fig:meta_action_general}
    \vspace{-1mm}
\end{wrapfigure}
we collectively refer to as its end-effector status. Thus, a robot's inherent actions can be abstracted as {move/rotate, ee status change}. However, to address diverse tasks, these inherent actions need to be contextualized within the environment. Therefore, we propose the following abstract meta-action to structure the planner's output, as shown in Eq. \ref{eq:1}. An example of how these meta-actions function as the planned result is shown in Fig. \ref{fig:meta_action}. Moreover, consider a single-arm robot with a gripper as the end-effector; in Fig. \ref{fig:meta_action_general}, we demonstrate that meta-actions can be combined in various ways to achieve higher-level sub-tasks that correspond to human concepts, ultimately offering greater flexibility and generality for solving daily activity tasks. Please see \ref{discussion_1} for further discussion about the design intentions of the meta-action. The execution of these meta-actions requires only a few execution functions, as illustrated in \ref{meta_action}.

\begin{figure}
    \vspace{+2mm}
\includegraphics[width=0.95\linewidth, height=35mm]{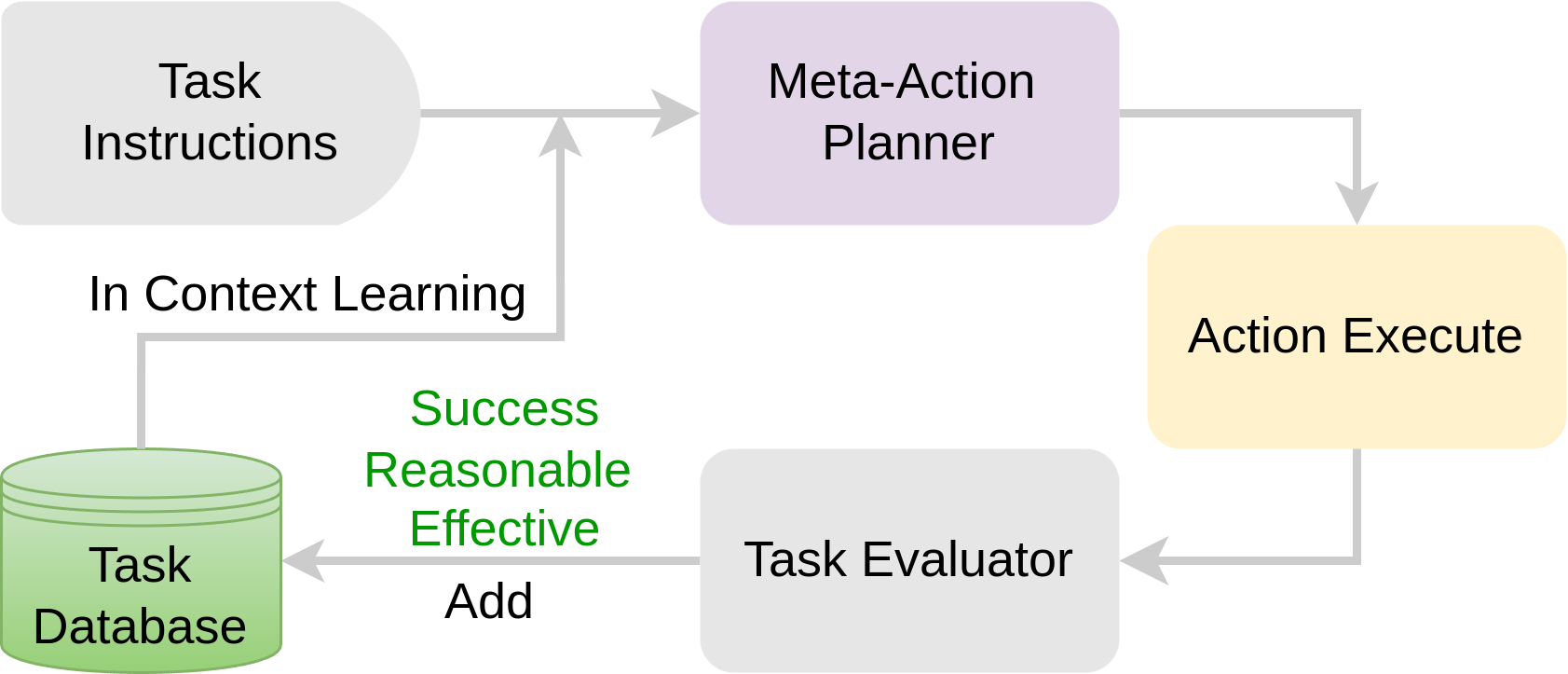}
    \caption{This figure illustrates the interaction between the task database and the planner. The planner retrieves the most similar example for in-context learning, and successfully verified tasks are subsequently added to the database.}
    \label{fig:task_database}
    \vspace{-4mm}
\end{figure}

To ensure that the planned outcomes can be successfully executed using the corresponding execution functions, the outputs of the task planner must adhere to the structured decomposition outlined by the meta-action framework. Therefore, we further enhance our prompt engineering using Retrieval-Augmented Generation (RAG) techniques. Specifically, we construct a database of well-planned tasks conforming to our desired format. By retrieving the most similar task, we facilitate in-context learning for the planner. Crucially, this task database features self-augmentation capabilities, allowing newly completed and verified tasks to be automatically incorporated into the database. This process progressively enhances the system's robustness. How the task database interacts with the meta-action planner is shown in Fig. \ref{fig:task_database}.

\textbf{Contributions} This paper presents a novel task planner, along with its corresponding execution actions. We summarize our contributions as follows:

\begin{itemize}    
    
    \item We propose MaP-AVR, a meta-action planner for agents that integrate Vision-Language Models (VLMs) and Retrieval-Augmented Generation (RAG). By abstracting skills into meta-actions and utilizing a self-augmented database with RAG, we establish a novel paradigm for task decomposition, ensuring the scalability and generalization ability of the proposed task planner.

    \item By utilizing the proposed task planner in conjunction with our defined successive execution functions, the proposed solution, as a type of agent-style method, surpasses the performance of recently published state-of-the-art agent methods. This demonstrates its potential for application across various domains, which are further discussed in \ref{discussion_2}.
        
\end{itemize}

\begin{figure*}[htbp]
   \vspace{+2mm}
    \centering
    \includegraphics[width=\textwidth]{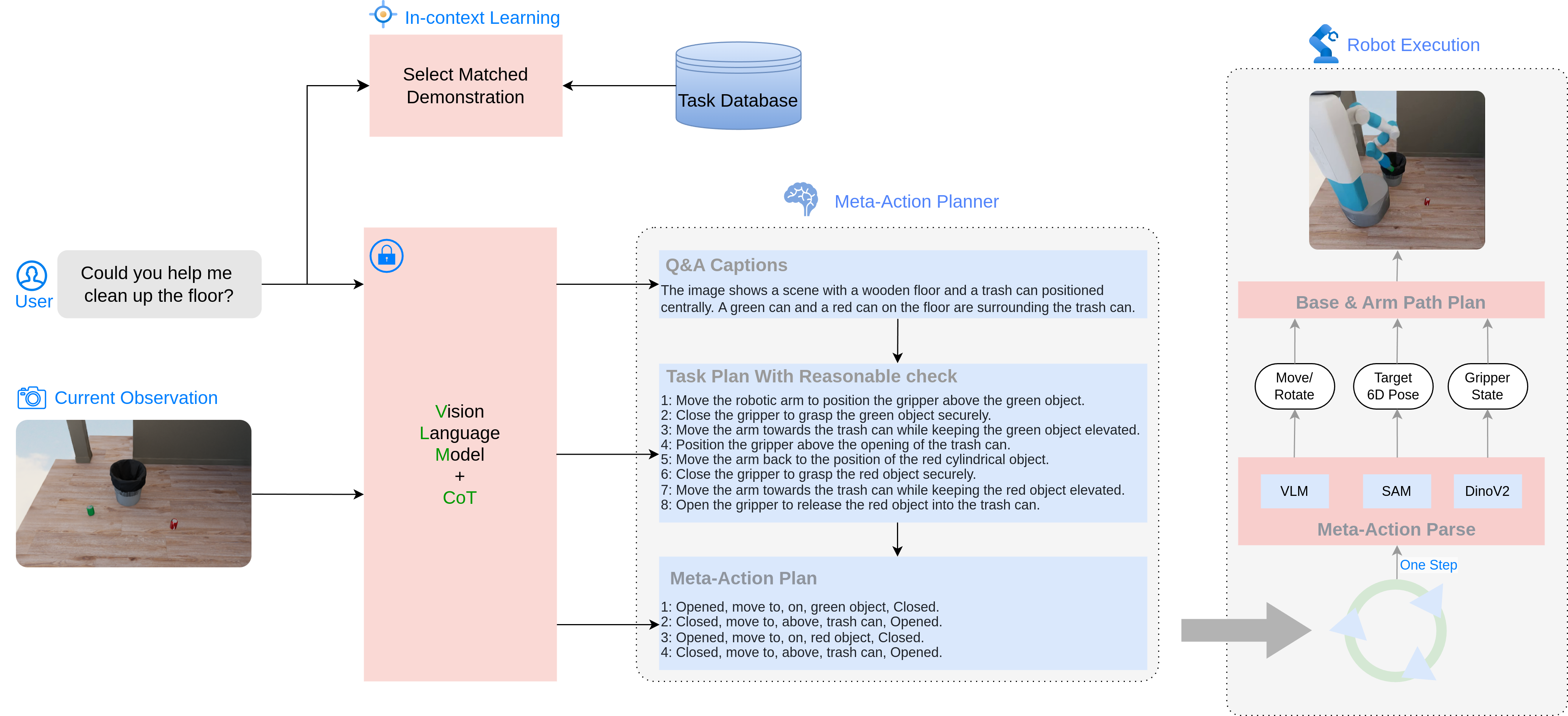}
    \caption{The overview of our proposed MaP-AVR. Compared to previous methods, our pipeline differs in each of its major components. \textbf{In understanding the task}, we incorporate the Retrieval-Augmented Generation (RAG) technique to search the database for the closest successful task to facilitate in-context learning. \textbf{In planning the task}, we designed a series of Chain-of-Thought (CoT) prompts to guide the model generating meta-actions that align with expectations. \textbf{In executing the task}, we leverage the spatial understanding ability of VLMs, in conjunction with foundation models(such as Sam, DinoV2) and classical obstacle avoidance algorithms, to ensure that the planned meta-actions are executed effectively and successfully.}
    \label{fig:wide_image}
    \vspace{-4mm}
\end{figure*}

\section{RELATED WORK}

\textbf{Single Task Methods} Early reinforcement learning and imitation learning methods typically focused on learning within a single environment or under the guidance of a single type of demonstration \cite{chi2023diffusion, ke20243d, jang2022bc, kalashnikov2021mt, zhao2023learning}. These methods have shown notable success in non-human interactive areas, such as motion control \cite{ma2024eureka}.

\textbf{Multi-action Models} To handle complex, everyday tasks, robots require the capacity to perform diverse actions. Recent advancements in multimodal language models have facilitated the incorporation of language embeddings as a common method to integrate interactive information \cite{kim2022diffusionclip}. Consequently, some research has explored introducing language signals into imitation learning frameworks, enabling neural networks to execute different actions based on language instructions \cite{goyal2024rvt, lynch2023interactive, ke2024d}. Some studies have gone further by using token replacement techniques to transform multimodal vision-language models (VLMs) into multimodal vision-language-action models \cite{brohan2022rt, brohan2023rt, belkhale2024rt}. This transformation imbues action models with the inherent world understanding acquired by VLMs during training on massive datasets.

\textbf{Task Planner in Two-Step Approaches} Despite incorporating language information into action networks, some recent researchers favor a two-step approach \cite{wake2024gpt, mu2024embodiedgpt, driess2023palm, lynch2023interactive, hu2024look, huang2024copa}: first, employing LLMs/VLMs to comprehend task descriptions and decompose them into multi-step action plans; and second, executing these plans using appropriate action models. However, a critical gap exists in the literature on task planning: a lack of thorough discussion regarding the optimal construction of action or skill sets. Most existing work relies on heuristic approaches, defining actions or skills based on human intuition and understanding, such as pick, move, place, open, close, etc.

\textbf{VLM Powered Robotic Agent} Beyond reinforcement learning, imitation learning, and vision-language-action models discussed above, AI agent-based approaches have emerged as a compelling avenue for developing robotic solutions, attracting considerable attention from researchers. With a VLM as its core, the system orchestrates a suite of foundation models (e.g., Grounding-Dino \cite{liu2023grounding} and GraspNet \cite{fang2020graspnet}) and traditional methods (e.g., obstacle avoidance), forming a training-free AI agent system capable of completing diverse everyday tasks \cite{liu2024moka, huang2024rekep, tziafas2024towards, duan2024manipulateanything}. In this paper, our definition of action execution functions draws inspiration from the implementation strategies employed in these agent-based approaches.

\textbf{CoT, ICL and RAG in Robotic} Recent advancements in large language models have introduced numerous powerful techniques that facilitate the emergence of advanced capabilities in these models, such as Chain-of-Thought (CoT) prompting \cite{wei2022chain}, in-context learning (ICL) \cite{brown2020language}, and Retrieval-Augmented Generation (RAG) \cite{lewis2020retrieval}. Studies consistently demonstrate that these methods unlock a remarkably high expressive capacity in large language models \cite{li2024chain}. Consequently, research exploring the application of these techniques to enhance robotic problem-solving is gaining significant traction. The use of Chain-of-Thought (CoT) prompting enables LLMs to perform direct path planning \cite{kwon2024language} and significantly enhances the capabilities of vision-language-action (VLA) models \cite{zawalski2024robotic}. Several studies have demonstrated the potential of in-context learning (ICL) techniques, even for high-precision tasks like grasping \cite{palo2024keypoint, yin2024context}. In contrast to CoT and ICL, the application of Retrieval-Augmented Generation (RAG) in robotics remains relatively under-explored. While recent work has demonstrated the use of RAG for robot navigation \cite{anwar2024remembr}, this paper represents an early exploration of its potential in enabling robots to handle everyday tasks.

\section{Methods}

\subsection{Preliminary}

Note that throughout the remainder of the paper, we use the following abbreviations: VLMs refer to vision-language models, RAG refers to Retrieval-Augmented Generation, ICL refers to in-context learning, and CoT refers to Chain-of-Thought. The VLM used in this paper is GPT-4o. In recent papers, VLM, RAG, ICL, CoT, and various foundation models have been combined in different forms as one or more components of robotic systems. However, the use of these techniques in this paper differs from previous works in the following ways:

In contrast to using RAG for robot navigation \cite{anwar2024remembr}, we build the RAG database to assist the planner in producing the desired outcomes. ICL is used as part of RAG, which contrasts with some studies where ICL is employed to guide VLM as a pattern generator that directly determines the position of the robotic gripper \cite{palo2024keypoint, yin2024context}. We design the CoT prompt from scratch to facilitate the alignment of the VLM's outputs with the meta-action format. As for foundation models, we do not extensively utilize models such as GraspNet \cite{fang2020graspnet}; instead, our method primarily focuses on unlocking the spatial understanding capabilities demonstrated in PIVOT \cite{nasiriany2024pivot}.


\subsection{Method Pipeline}

We illustrate the entire pipeline of the proposed MaP-AVR in Fig. \ref{fig:wide_image}. When the robot receives an instruction for a daily task, it first searches its existing experience database for the most similar task and scenario. The located completed task, along with its prompting cache, is used as a demonstration and is sent to the VLM together with the task instruction. We design a series of prompts, such as having the VLM describe the scenario and identify task-related objects, among others. This process results in a collection of step-by-step actions that are described in natural language and can be verified by the VLM. After a reasonableness check, we generate a series of meta-actions as the final planning result. These meta-actions are sent to the corresponding execution functions sequentially. 


\subsection{Meta-Action Set} \label{meta_action}


\subsubsection{The definition of the meta-action set}

\begin{figure}
    \vspace{+2mm}
\includegraphics[width=1.\linewidth, height=25mm]{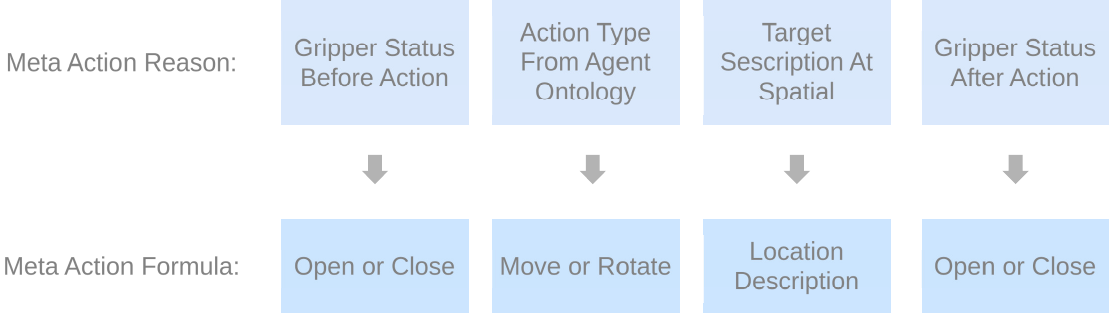}
    \caption{This figure shows the linguistic structure of the meta-action we finally guide the VLM to generate.}
    \label{fig:meta-action-set}
    \vspace{-4mm}
\end{figure}

The meta-action is proposed to achieve two objectives: first, to maintain the generality of the robotic system, and second, to ensure executability by originating actions from the robot's actuators. Based on the design principle in \ref{discussion_1}, we define the format of each instruction set as follows: gripper status before action (Open or Close), move or rotate, location description such as the direction (e.g., left/right/up/down/forward/backward) of the target object, and gripper status after reaching the target pose. For example, the meta-action could be “opened, move to, front on, burger, closed.” In Fig. \ref{fig:meta-action-set}, we illustrate this transition process that guides the VLM to output meta-actions in the desired fixed linguistic structure. The first and last elements describe the status change of the end-effector. The second element involves the basic action of the end-effector, such as moving or rotating. The middle part depicts the final relationship between the end-effector and the environment after the action has been executed, thereby providing a textual goal for the current action. These three parts are abstracted in Eq. \ref{eq:1}.

\vspace{-4mm}
\begin{equation}\label{eq:1}
\{move/rotate, \hspace{1mm}location \hspace{1mm}description, \hspace{1mm} ee\hspace{1mm} status\}.
\end{equation}

\subsubsection{The CoT prompt for the meta-action set}
\begin{figure}
    \vspace{+2mm}
\includegraphics[width=1.\linewidth, height=70mm]{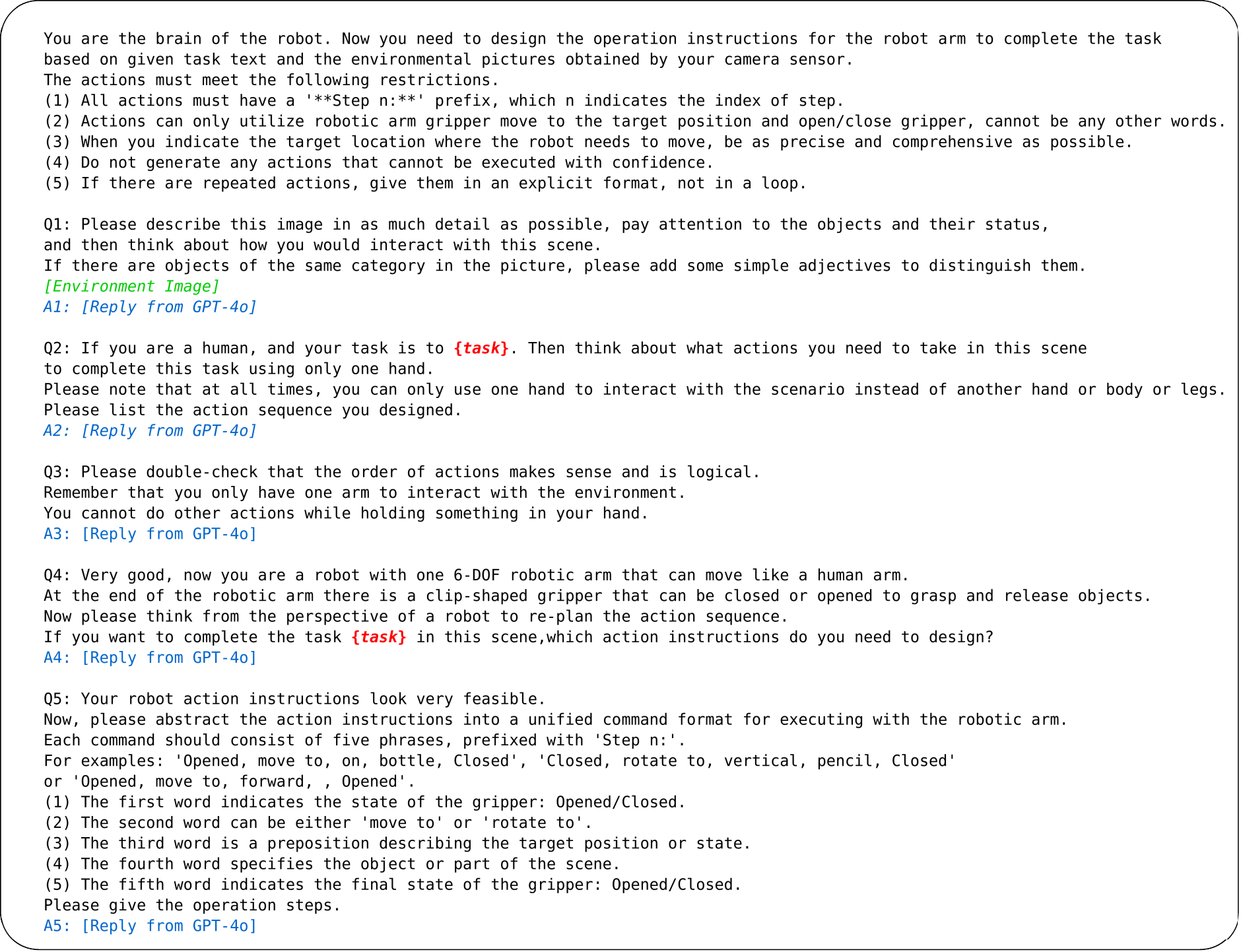}
    \caption{This figure illustrates the prompts used for meta-action generation.}
    \label{fig:meta-action-cot}
    \vspace{-4mm}
\end{figure}

As shown in Figure \ref{fig:meta-action-cot}, we designed a multi-turn conversation that includes a prompt system based on the CoT principle to help the VLM generate meta-action results. This conversation reflects an understanding of the scenario and task. It leverages the VLM's general knowledge capabilities to produce a coherent task planning sequence. The prompts explicitly guide the VLM to optimize the task planning sequence for rationality and correctness. Finally, the definition of the meta-action is provided, and the VLM is instructed to output the final task planning sequence in the meta-action format based on its understanding.

\subsubsection{Move/Rotate and Gripper executor implementation} \label{action_paragraph}

As illustrated in Figures \ref{fig:wide_image} and \ref{fig:meta-action-set}, the meta-actions are carried out through corresponding functions. The robot utilized in this study is a single-arm Fetch Robot, therefore, we refer to the end effector simply as the gripper. The gripper execution function follows a straightforward logic: meta-actions such as \{Open,..., Close\} and \{Close,..., Open\} will operate the gripper by closing and opening it, respectively. In contrast, sequences like \{Open,..., Open\} and \{Close,..., Close\} will maintain the current status of the gripper. The key factor in the meta-action is the $location \hspace{1mm} description$ term, which serves as the textual goal for each meta-action. While executing each meta-action, the $location \hspace{1mm} description$ will be finally transformed to one specific executable target 6-Dof pose denoted as $P_{target}$. The transformation process roughly follows the process below:

\begin{itemize}    
    
    \item Step 1, utilize VLM to find an initial 3D point along with a default orientation as the $P_{init}$ via visual prompting, shown in Eq. \ref{eq:2}. If the $location \hspace{1mm} description$ contains only a preposition, use the last position as the initial 3D point.
    
\vspace{-4mm}
\begin{align}\label{eq:2}
VLM(location \hspace{1mm} description) \longrightarrow P_{init}
\end{align}

    \item Step 2, uniformly sample $n$ offset 6-Dof poses to prepare $n$ candidates, shown in Eq. \ref{eq:3}. 
    When the move action is executed with a single preposition as the $location \hspace{1mm} description$, the offset can only sample translation. For the rotate action, the offset can only sample rotation. 

\vspace{-4mm}
\begin{align}\label{eq:3}
P_{candidate}^{i} = P_{init} + P_{offset}^{i} 
\end{align}

    \item Step 3, the VLM will be used as a selector to choose the best candidate that aligns most closely with the overall goal, shown in Eq. \ref{eq:4}.
    
\vspace{-4mm}
\begin{align}\label{eq:4}
P_{target} = Select(\{P_{candidate}^{i}, i \in [1,...,n]\})
\end{align}

\end{itemize}

\begin{figure}
    \vspace{+2mm}
    \centering
    \includegraphics[width=0.99\linewidth, height=60mm]{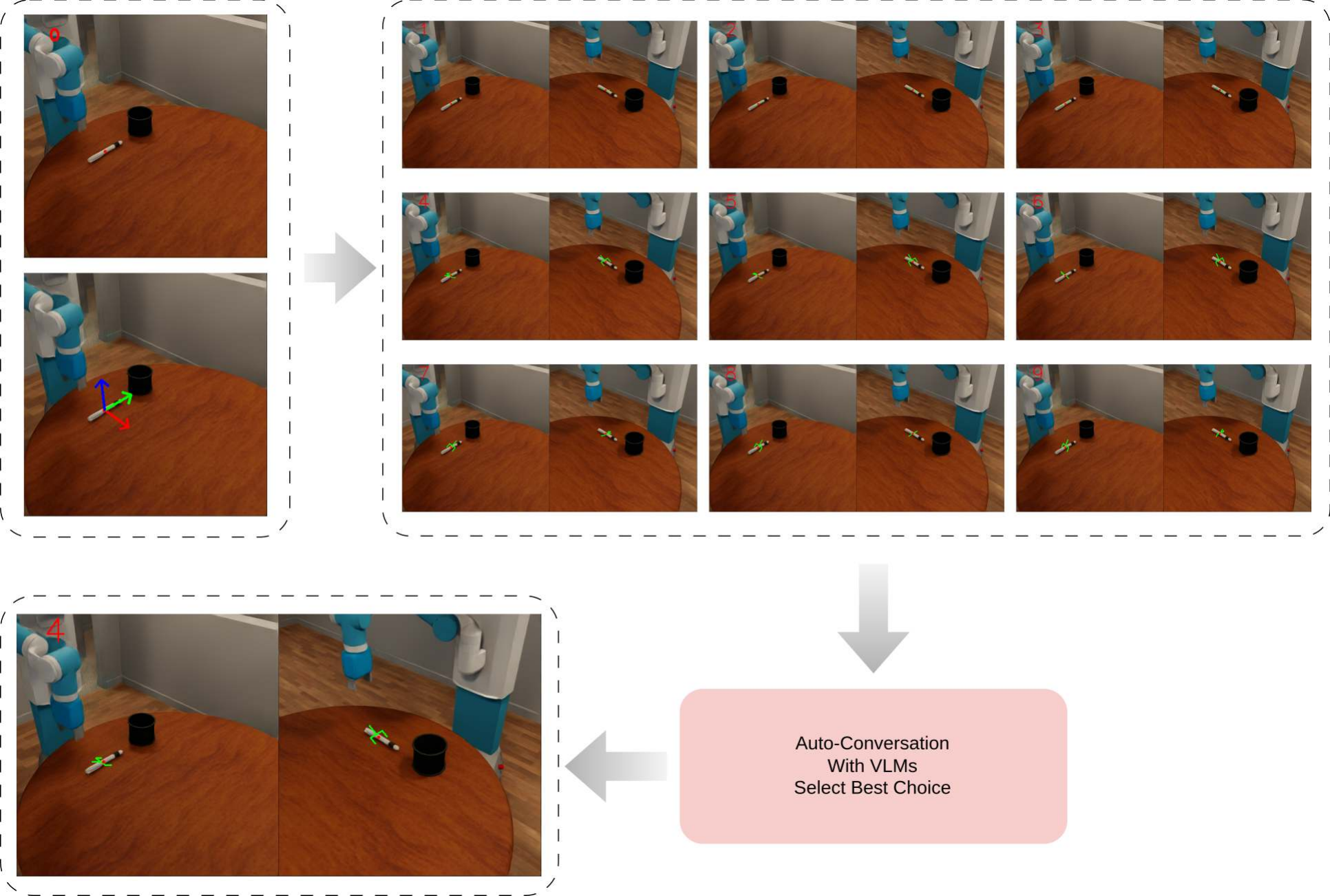}
    \caption{This figure illustrates how we uniformly sample multiple candidate poses around the initial pose and rely on the VLM to determine the final target pose.}
    \label{fig:candidate_chosen}
    \vspace{-4mm}
\end{figure}

Instead of directly using foundation grasping models like GraspNet, the action execution functions make the greatest possible use of the general knowledge embedded within the VLM, as highlighted by PIVOT \cite{nasiriany2024pivot}. This specific candidate selection process is visually shown in Fig. \ref{fig:candidate_chosen}. After the arm reaches the target pose, the gripper action is executed.

    


    
    

\subsection{RAG in Robotic Task Planning }

\subsubsection{Why the RAG is needed in our task planning} The definition of meta-action requires the output of the VLM to maintain a specific format. As shown in Fig. \ref{fig:meta_action}, the first element of the $ee \hspace{1mm} status$ should be the same as the last element in the previous meta-action. For the $location \hspace{1mm} description$, the use of prepositions is crucial. For example, ``above'' signifies moving to a position over the object while maintaining a certain distance, whereas ``on" denotes directly interacting with the object. Utilizing RAG technology can help eliminate the potential confusion that may arise from pre-trained VLMs. Moreover, we empirically found that replacing the most similar examples with less similar ones indeed caused some previously successful plans to fail. This aligns with the anticipated benefits of utilizing RAG and ICL.

\subsubsection{Database preparation and planner performance evaluation via human annotating} \label{eval_rag} To utilize RAG, we need to prepare a database of successful planning demonstrations. Those initial demonstrations in this database relying on human involvement. Additionally, since our planner and execution functions are decoupled, the planner can be used to generate plans for any scenario independently. However, evaluating the planner's performance on arbitrary data without relying on the executor also necessitates human involvement. Inspired by \cite{wake2024gpt}, we have developed multiple web interfaces connected to the database to support these human-involved functions. These interfaces facilitate the invocation of the planner, modifications to the entire intermediate planning process, saving of correct planning results, comparisons of planning outcomes for arbitrary images with and without ICL, and manual voting on the correctness of the planning results, among other functionalities.

\subsubsection{The design of the RAG integration} With the initial database in place, we can begin querying and augmenting the planner using RAG. For the retrieval component, we utilize VLMs to extract the object scene graph from the image, serving as a scene description. This scene graph, along with the task instruction, is used to embed the task into a suitable representation. When a new task arises, we query the top-k samples and utilize VLMs to select the most similar one. For the augmentation component, we retrieve the saved GPT-cache of the selected demonstration, which contains all the prompts and replies used in the demonstration's planning process, and add the GPT-cache as an additional dialogue round.

\subsubsection{The extended of the RAG database}
For the RAG database, In addition to human involvement,  it can also grow autonomously as more tasks are completed. We evaluated two metrics for every task in the database with method based VLM (GPT-4o). That are the similarity of relevant objects and the similarity of sequences from the task planner normalized to $[0, 1]$. If a new task is executed correctly and these two metrics calculated across all tasks in the database are lower than preset, suggesting that the new task has low cross-correlation with the all existing task in the database, the new task will be added to the task database. The growth of the RAG database will enhance the meta-action planner to predict correctly result for other tasks that has low cross-correlation with the RAG database.

\section{EXPERIMENTS}

\subsection{Datasets}

\subsubsection{OmniGibson} 
\begin{figure}
    \vspace{+2mm}
    \centering
    \includegraphics[width=1.\linewidth, height=60mm]{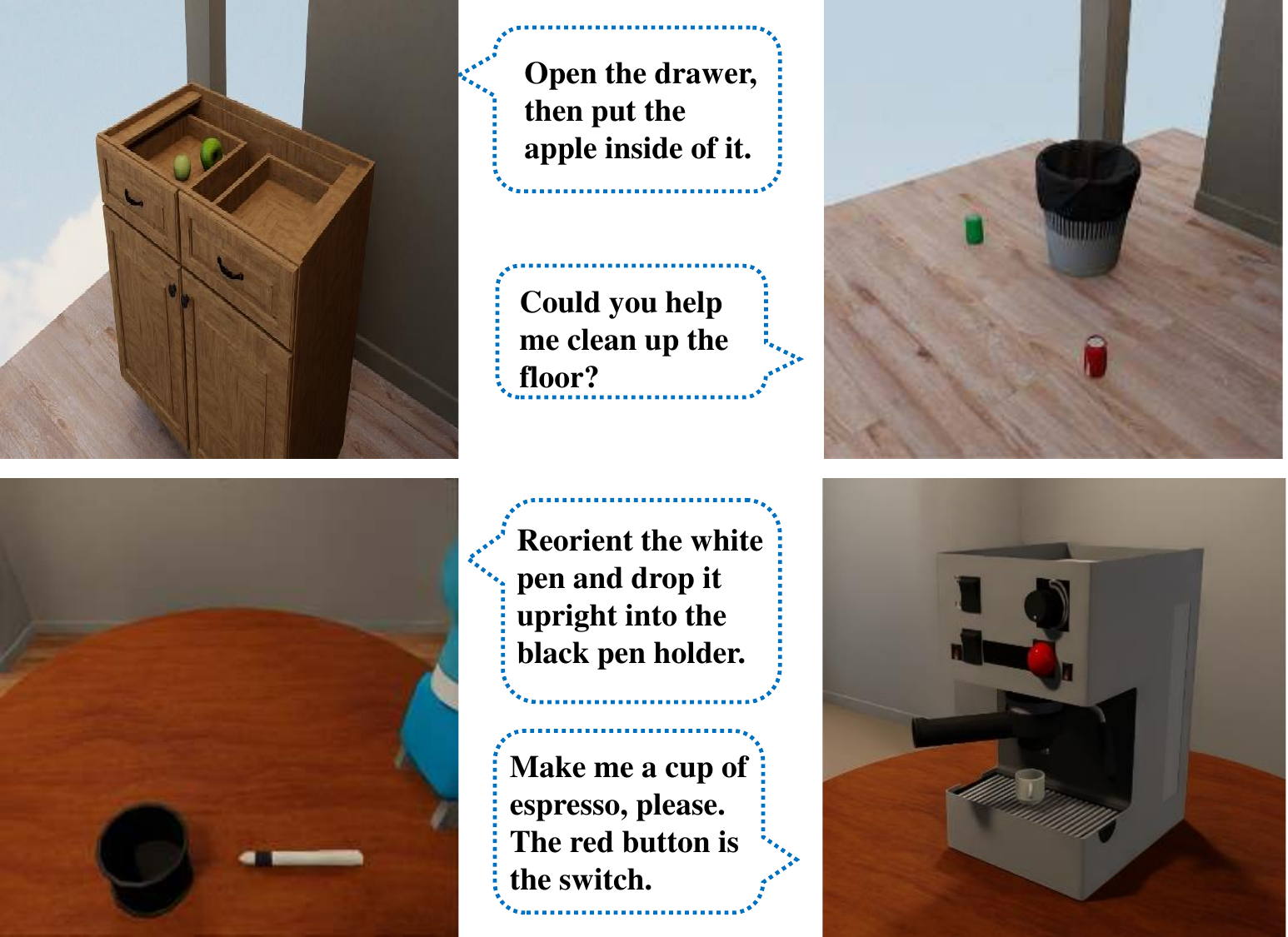}
    \caption{This figure shows some daily tasks that we developed in OmniGibson to test our method.}
    \label{fig:omnigibson}
    \vspace{-4mm}
\end{figure}

Identifying suitable datasets for evaluating embodied intelligence has long been a challenging issue in the field. While robotic data gathered from real-world operations tend to be more realistic and credible, such data is often closely tied to the specific robot from which it was collected, making the methods developed on these datasets difficult to reproduce widely. Testing methods in simulators are reproducible, but the tasks offered by most simulators are relatively simple compared to the complex tasks encountered in daily life. Therefore, to address this issue, we choose to use OmniGibson as our simulator. Built upon the powerful NVIDIA Omniverse platform, it allows users to develop and define their own scenes and tasks. We develop several scenarios and tasks according to our own needs, as shown in Fig. \ref{fig:omnigibson}.

\subsubsection{The mixed image collection} \label{mix_dataset} 
\begin{figure}
    \vspace{+2mm}
    \centering
    \includegraphics[width=0.99\linewidth]{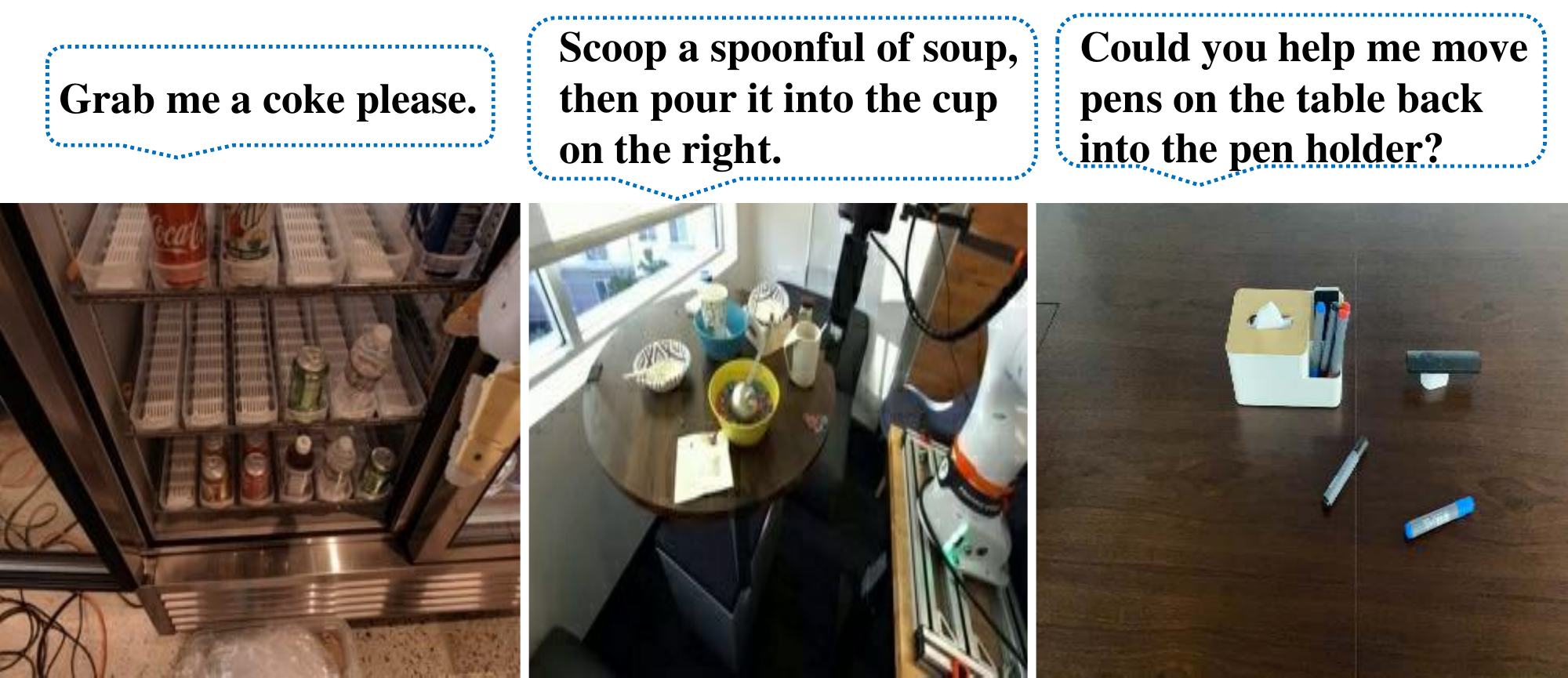}
    \caption{This figure displays the images from our mixed data collection, which are used to evaluate the meta-action planner.}
    \label{fig:mix_collection}
    \vspace{-4mm}
\end{figure}

As mentioned in \ref{eval_rag}, the planner and subsequent execution functions are decoupled. For evaluating the task planner, we can perform planning for any task using any image and subsequently assess the results through manual voting by humans without involving the execution component. We prepared a mixed data collection by sampling images from RT-1 \cite{brohan2022rt}, RoboVQA \cite{sermanet2024robovqa}, and Droid-100 \cite{khazatsky2024droid} sourced from the Open X-Embodiment data pool \cite{o2023open}. Additionally, we captured several office scenes, which, together with the previously sampled images, formed the mixed collection used in this study. Some samples are shown in Fig. \ref{fig:mix_collection}. Note that all task instructions were randomly provided by human annotators based on the images, ensuring that each image has a unique task.

\subsection{Performance Comparison with the State-of-the-art}
\begin{table}[h]
   \vspace{-1mm}
  \begin{center}
    \caption{Performance Comparison with Rekep on OmniGibson}
    \label{tab:exp_omnigibson}
    \setlength{\tabcolsep}{6pt} 
    \renewcommand{\arraystretch}{1.2} 
    \begin{tabular}{c|c|c|c|c}
      \hline
       Task Type &  n-trials & Rekep & Ours w/o ICL & Ours w ICL    \\
       \hline
       Insert the pen     & 40 &  14/40 & 	4/40  &  18/40 \\
       Clean up the floor & 40 &  8/40  & 	3/40  &  33/40 \\
       Open drawer        & 40 &  0/40  & 	8/40 &  10/40  \\
       Make coffee        & 40 &  0/40  & 	3/40 &  8/40 \\
              \hline
        success rate & 160 & 13.75\% & 11.25\% & 43.13\%  \\
              \hline
    \end{tabular}
  \end{center}
   \vspace{-4mm}
\end{table}

We set the benchmark by selecting Rekep \cite{huang2024rekep} as the baseline state-of-the-art comparison method. Rekep is a recently published work that has achieved significant results compared to previous methods. The authors have open-sourced their code, utilizing OmniGibson as the demonstration platform. This allows us to directly use their open-source code, thereby avoiding the uncertainty and risks associated with re-implementing the algorithm.

The comparison results are shown in Table \ref{tab:exp_omnigibson}. The method proposed in this study significantly outperforms Rekep. Although Rekep surpasses previous methods such as Voxposer \cite{huang2023voxposer}, its success rate is relatively limited and only increases significantly when the required feature points are manually provided, as reported in their paper. In contrast, our method achieves higher success rates without the need for human intervention, demonstrating a stronger ability to solve complex tasks.

\subsection{Ablation Study}

\subsubsection{Failure case analysis}

\begin{wrapfigure}{r}{.6\linewidth}
     \vspace{-2mm}
        \includegraphics[width=1.\linewidth]{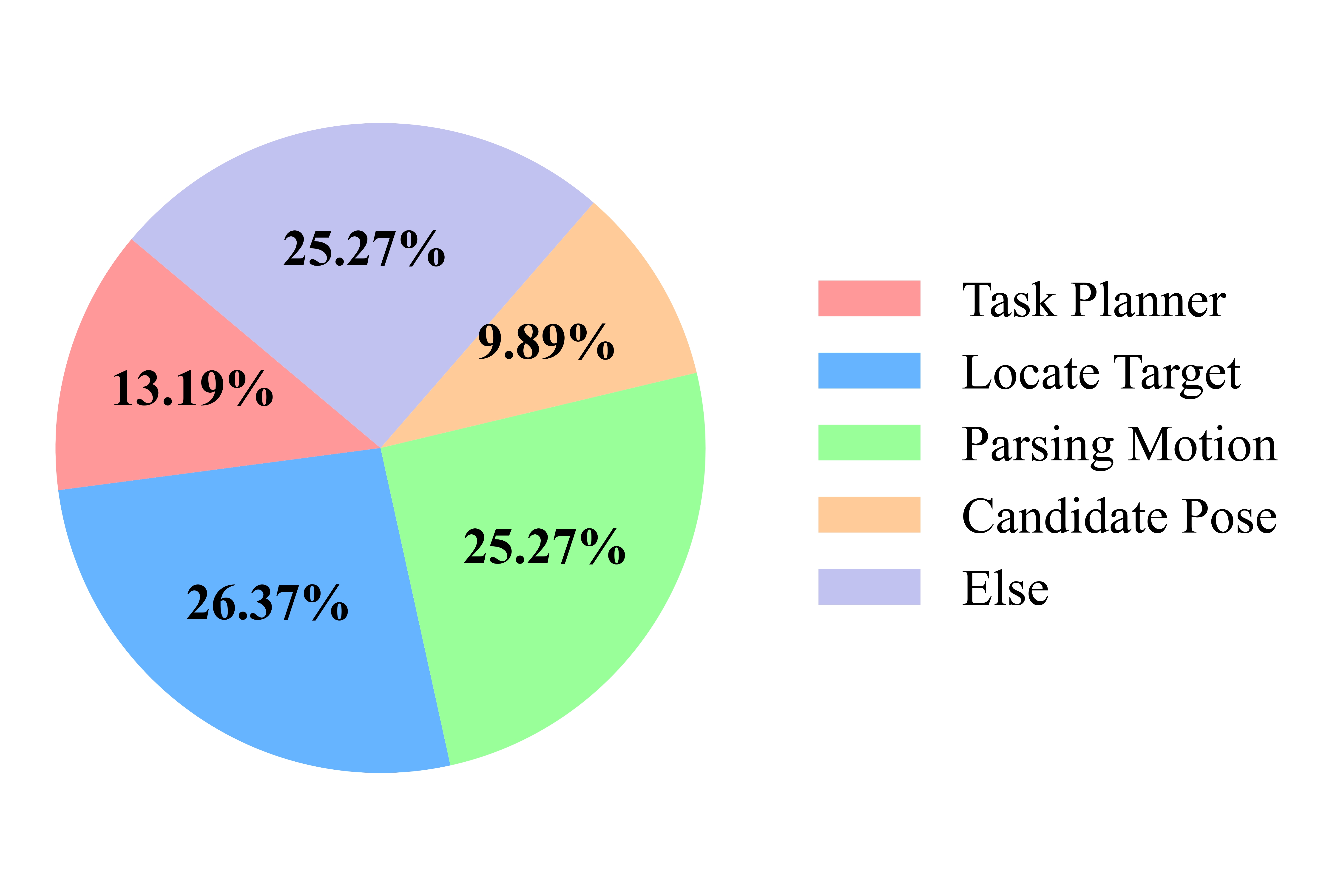}     
     \vspace{-5mm}
    \caption{Statistical analysis of the primary failure reason for each trial.}\label{fig:failure}
    \vspace{-3mm}
\end{wrapfigure}

 Here, we further analyze which steps contribute to the final failure in Fig. \ref{fig:failure}. It can be seen from the figure that the failure to locate the target object and the target grasping point is the main reason (about 26\%), which is usually the basic input of the subsequent VLM. Therefore, the accuracy and rationality of the
 target point positioning directly affects the judgment of the subsequent VLM. Especially for the `open drawer' and `make coffee' tasks, it is difficult to find the target points of the red button of the
coffee machine and the handle of the drawer. The second is the parsing action, accounting for about 25\%, which is mainly manifested in the error in judging the rotation axis when parsing the rotation action through VLM. In addition, the proportion of cases caused by unreasonable task planning and candidate pose generation is relatively small, about 13\% and 10\% respectively.

\subsubsection{How much the RAG can help} 

Using Retrieval-Augmented Generation (RAG) enables us to perform in-context learning. Although the reasons for task failure can be diverse, a better planner should theoretically be able to statistically improve the success rate of task completion. In Table \ref{tab:exp_omnigibson}, we compare the success rates on OmniGibson with and without in-context learning (ICL). We observe that the planner equipped with ICL dramatically improves the overall task success rate. 

In addition to indirectly evaluating task success rates, we prepared a diverse mixed image dataset specifically designed to test the task planner (see Fig. \ref{mix_dataset}). Using the web interface we developed, human annotators can assign arbitrary task instructions based on the images they select. These tasks can vary in complexity, being either simple or complex, depending on the annotators' discretion. The back-end calls our planner and returns the planned results, both with and without ICL. Human annotators then assess whether, if they were the robot, they could complete the task by following the planned result. They click the corresponding buttons to indicate whether the planned results are correct or not. The planning success rate is shown in Table \ref{tab:plan_eval}, demonstrating that ICL significantly improves planning. We also discovered that in the absence of ICL, the meta-action planner exhibits certain consistent error patterns, such as providing redundant actions or confusing directions during push and pull tasks.

\begin{table}[h]
   \vspace{-1mm}
  \begin{center}
    \caption{Performance Comparison without and with ICL on the Mixed Data Collection.}
    \label{tab:plan_eval}
    \setlength{\tabcolsep}{6pt} 
    \renewcommand{\arraystretch}{1.2} 
    \begin{tabular}{c|c|c|c}
      \hline
       Data Source &  n-tasks & w/o ICL & with ICL     \\
       \hline
       RT-1         &  20 & 6/20  &  13/20  \\
       RoboVQA      &  20 & 4/20  &  13/20  \\
       Droid-100    &  20 & 8/20  &  14/20  \\
       Office       &  50 & 17/50 &  39/50  \\
       \hline
       success rate &  110& 31.8\% &   71.8\%  \\
       \hline
    \end{tabular}
  \end{center}
   \vspace{-4mm}
\end{table}

\section{Discussion} 

\subsection{The Design Principle of the Meta-Action} \label{discussion_1}
We believe an effective task planner should exhibit both generalization ability and comprehensiveness. To enhance generalization ability, we abstract skills into meta-actions encompassing only ``move", ``rotate", and end-effector actions. These are the fundamental commands that the robot is inherently capable of executing. However, this abstraction introduces challenges in maintaining comprehensiveness across all scenarios. To address the challenge of maintaining comprehensiveness while abstracting skills into meta-actions, we draw inspiration from John Maynard Keynes's famous adage: ``It is better to be roughly right than precisely wrong". Specifically, we introduce a $location \hspace{1mm} description$, which provides a more flexible and approximate representation of the action's end status. Thus, our meta-action does not represent a rigid action command but rather a description of the intended action and its desired outcome. This approach ensures that our planner is ``roughly right" at the task planning level. This is crucial because an incorrect plan at this stage would render the entire system incapable of completing the task. The process of translating this ``roughly right" plan into a ``precisely right" action command is then deferred to the action execution functions.

\subsection{Integrate the Proposed Planner with Other Methods} \label{discussion_2}
The proposed task planner, in conjunction with our defined execution functions, constitutes an AI agent-based solution for robotic daily task completion. This solution can independently solve problems like similar agent-style methods VoxPoser \cite{huang2023voxposer}, VILA \cite{hu2024look} + CoPa \cite{huang2024copa}, MOKA \cite{liu2024moka}, and Rekep \cite{huang2024rekep}.

\textbf{Generating Data for Imitation Learning} A recent study introduced an agent-style method and demonstrated that using it to collect demonstrations for imitation learning network training could produce results comparable to those obtained from human-provided demonstrations \cite{duan2024manipulateanything}. The method presented in this paper exhibits the capability to achieve a similar outcome.

\textbf{Working with the Policy Network} In many research papers, the action executor takes the form of a policy network \cite{lynch2023interactive, ke2024d}. Unlike most other agent-style methods that understand tasks as visual prompts coupled with successive execution \cite{liu2024moka, huang2024rekep}, the task planner proposed in this paper can function independently, utilizing a policy network as its action executor.

\section{CONCLUSIONS}
Decomposing human instructions into executable signals through a task planner is a crucial step toward advancing robotic intelligence. In this letter, we propose MaP-AVR, a meta-action planner for agents composed of Vision-Language Models (VLMs) and Retrieval-Augmented Generation (RAG), to explore this direction. We define a general-purpose meta-action to guide the multimodal large language model in formatting its outputs into directly executable instructions. Retrieval-Augmented Generation (RAG) is utilized to consolidate the planning process, ensuring better alignment with our desired format. The experiments were conducted in robotic simulation environments developed with OmniGibson, ensuring that all results are reproducible and extendable. We believe our research benefits the community by enhancing how embodied AI can better understand and perform human daily activities.












\bibliographystyle{IEEEtran}
\bibliography{IEEEabrv,IEEEexample}

\end{document}